\documentclass[conference]{IEEEtran}
\IEEEoverridecommandlockouts
\usepackage{cite}
\usepackage{amsmath,amssymb,amsfonts}
\usepackage{algorithmic}
\usepackage{graphicx}
\usepackage{textcomp}
\usepackage{xcolor}
\usepackage{multirow}
\usepackage{utfsym}
\def\BibTeX{{\rm B\kern-.05em{\sc i\kern-.025em b}\kern-.08em
    T\kern-.1667em\lower.7ex\hbox{E}\kern-.125emX}}
\begin{document}

\title{Spatial Parsing and Dynamic Temporal Pooling networks for Human-Object Interaction detection\\
\thanks{This work was supported by the National Key Research and Development Program of China under Grant 2020YFF0304900 and National Natural Science Foundation of China (62072358,62073252) and Chinese Defense Advance Research Program (50912020105).}
}

\author{
\IEEEauthorblockN{
Hongsheng Li\IEEEauthorrefmark{1},
Guangming Zhu\IEEEauthorrefmark{1},
Wu Zhen\IEEEauthorrefmark{2}, and
Lan Ni\IEEEauthorrefmark{3}
Peiyi Shen\IEEEauthorrefmark{1}
Liang Zhang\IEEEauthorrefmark{1}
Ning Wang\IEEEauthorrefmark{1}
Cong Hua\IEEEauthorrefmark{1}}
\IEEEauthorblockA{\IEEEauthorrefmark{1}School of Computer Science and Technology, Xidian University, China}
\IEEEauthorblockA{\IEEEauthorrefmark{2}Beijing Roborock Technology Co.,Ltd.}
\IEEEauthorblockA{\IEEEauthorrefmark{3}Shanghai University}
\IEEEauthorblockA{Corresponding Author: Liang Zhang \quad Email: liangzhang@xidian.edu.cn}}


\maketitle

\begin{abstract}

The key of Human-Object Interaction(HOI) recognition is to infer the relationship between human and objects. Recently, the image's Human-Object Interaction(HOI) detection has made significant progress. However, there is still room for improvement in video HOI detection performance. Existing one-stage methods use well-designed end-to-end networks to detect a video segment and directly predict an interaction. 
A side effect of these approaches is that we have no way of knowing the human-object pair of the interaction or the keyframes in which the interaction took place.
It makes the model learning and further optimization of the network more complex. This paper introduces the Spatial Parsing and Dynamic Temporal Pooling (SPDTP) network, which takes the entire video as a spatio-temporal graph with human and object nodes as input. Unlike existing methods, our proposed network predicts the difference between interactive and non-interactive pairs through explicit spatial parsing, and then performs interaction recognition. Moreover, we propose a learnable and differentiable Dynamic Temporal Module(DTM) to emphasize the keyframes of the video and suppress the redundant frame. Furthermore, the experimental results show that SPDTP can pay more attention to active human-object pairs and valid keyframes. Overall, we achieve state-of-the-art performance on CAD-120 dataset and Something-Else dataset.
\end{abstract}


\section{Introduction}
The task of Human-Object Interaction understanding aims to infer the relationships between humans and objects. For example, in a video of "drinking from a cup," we not only have to recognize that this is an activity of "drinking" but also localize the water cup being used from many objects that exist in the scene. HOI tasks involve object detection\cite{ren2015faster, wang2018repulsion}, object tracking\cite{minderer2019unsupervised, wang2019fast, bertinetto2016fully} and human pose estimation\cite{dabral2018learning, dabral2019multi, guler2018densepose}. The HOI research is significant to human-computer interaction, intelligent robots, visual question answering, etc.  

With the success of deep learning in various computer vision tasks, significant progress has been achieved in human-object interaction detection with various deep neural networks. Compared with image data, video data provides additional clues to identify objects in the interaction, but also increases the difficulty of interaction recognition. On the one hand, existing methods\cite{gupta2015visual, zhuang2017care, gao2018ican, qi2018learning, chao2018learning} take advantage of one-stage networks to detect and classify HOIs, simultaneously. They construct the video with many objects into a dense graph and directly predict all possible human-object pairs. In other words, they treat each object in the video equally, but in fact, not all objects contribute to the interaction behavior in the video, many objects in a video are irrelevant to the current interaction. The strength of the relationship between each object and human is different.  The human-object pairs formed in this way seriously affect the performance of the HOI recognition network. Meanwhile, in the process of relation mining, we cannot explicitly know the strength of the relationship between humans and objects. 

On the other hand, existing methods ignore the importance of the analysis on the temporal dimension. Furthermore, videos depict events about humans, such as their activities and behaviors, with a lot of redundant information in the time dimension. A straightforward method for temporal fusion in video processing is conducting temporal neural network(TCN) or recurrent neural networks(RNN). However, these methods treat all frames equally without discriminating them by importance. In fact, only a few keyframes are needed to express the entire interaction, the other frames are mostly background frames. 
Temporal fusion lies in parsing in time and emphasizing the compelling frames in videos that can represent the distinct interactions and suppress redundant frames. 

To address these problems, we proposed a Spatial Parsing and Dynamic Temporal Pooling network(SPDTP), which consists of five modules, among which the most important are \textbf{Spatial Parsing} and \textbf{Temporal Fusion}. We argue that HOI detection and human-object pairs parsing are tightly coupled and can support each other. \textbf{Spatial Parsing} enables the spatial subnet to automatically learn the strength of the human-object relationship based on various features of the nodes. These relationships are used to suppress non-interactive pairs and further reduce the influence of objects unrelated to interaction on interaction detection. 

The second is \textbf{Temporal Fusion} which is illustrated in Figure \ref{TP}. DTM is composed of a temporal convolution, equipping with Dynamic Temporal Pooling(DTP) instead of fixed-weights pooling. DTP is based on the principle that redundant frames in the video have high similarity, and frames can be clustered or abandoned according to the similarity and distinction so as to achieve the temporal parsing effect. Through the DTP, in the process of temporal fusion, we can simultaneously identify some keyframes, and remove the redundant information in the video.

Our proposed approach is evaluated on two challenging datasets, including the CAD-120, Something-Else datasets. Experimental results show that our model significantly outperforms state-of-the-art methods. In particular, our approach achieves 91.8\% and 90.2\% on subactivity and affordance F1 scores only using isolated videos on the CAD-120 dataset, respectively. And on the Something-Else dataset, our approach achieves 65.0\% and 89.2\% on top-1 and top-5 accuracy.

\section{Related work}

In recent years, the HoI detection problem has gradually become the focus of machine vision research. This section reviews previous work in HOI detection and graph neural networks.

\subsection{HOI detection}
This task starts from the idea of “affordances” introduced by J.J. Gibson\cite{gibson2014ecological}. It aims to identify not only the kind of interaction but also the objects that interact. HOI prediction represents the identification of all interactions that occurred within this video. Compared with image-level HOI prediction tasks, video has an extra dimension of time. Therefore how to use time cues to improve performance is the focus of current research. 

There have also been significant advances in HOI detection research in videos, mainly on the CAD-120 dataset. Koppula et al. \cite{koppula2013learning} proposed this dataset and introduced MRF-based formulations to deal with spatiotemporal requirements. The authors handcrafted a set of features for both humans (pose, joint displacement, etc.) and objects (3D centroids, SIFT matching transformations between adjacent frames, etc.). These features are not used at the frame level, but collectively represent an overall video segment. Similarly, Jain et al. \cite{jain2016structural} designed a spatiotemporal graph to perform structured prediction on videos composed of multiple segments. Chiou et al. \cite{chiou2021st} further mine the temporal information in videos and propose spatiotemporal HOI detection (ST-HOI), which utilizes temporal information such as human and object trajectories, correctly localized visual features, and spatiotemporal pose features.

\subsection{Graph neural network}
Some methods \cite{simonovsky2017dynamic, yuan2017temporal, defferrard2016convolutional} generalize neural network operations (such as convolution) directly from images to graphs.
However, the HOI problem requires a structured representation to capture the high-level spatiotemporal relationships between humans and objects. Some other works combine network architectures with graph models \cite{chen2017deeplab, chen2015learning} and have achieved results in applications such as object detection and parsing\cite{yuan2017temporal,liang2016semantic}, scene understanding\cite{li2017situation,marino2016more, xu2017scene} and VQA  \cite{teney2017graph}. However, these methods are only suitable for problems with a pre-fixed graph structure. Long short-term memory (LSTM) \cite{liang2017interpretable} is used to merge graph nodes to solve human parsing problems, assuming nodes are mergeable.
Qi et al. \cite{qi2018learning} proposed their GPNN method for videos and constructed a spatiotemporal graph. But it used the hand-craft features to initialize the node features and edge relations in their initial graph and neglected to deal with temporal features. Sunkesula et al. \cite{sunkesula2020lighten} introduced the end-to-end LIGHTEN network to detect human-object interactions from videos. In the process of spatial analysis, the network uses Graph Attention Networks(GAT)\cite{velivckovic2017graph} to automatically parse the strength of the relationship between objects. However, although LIGHTEN was able to parse the spatial graph by AGCN, this model only used one layer of RNN to parse the temporal graph in the time dimension.

\section{Method}

\begin{figure*}
	\includegraphics[width=500px,height=210px]{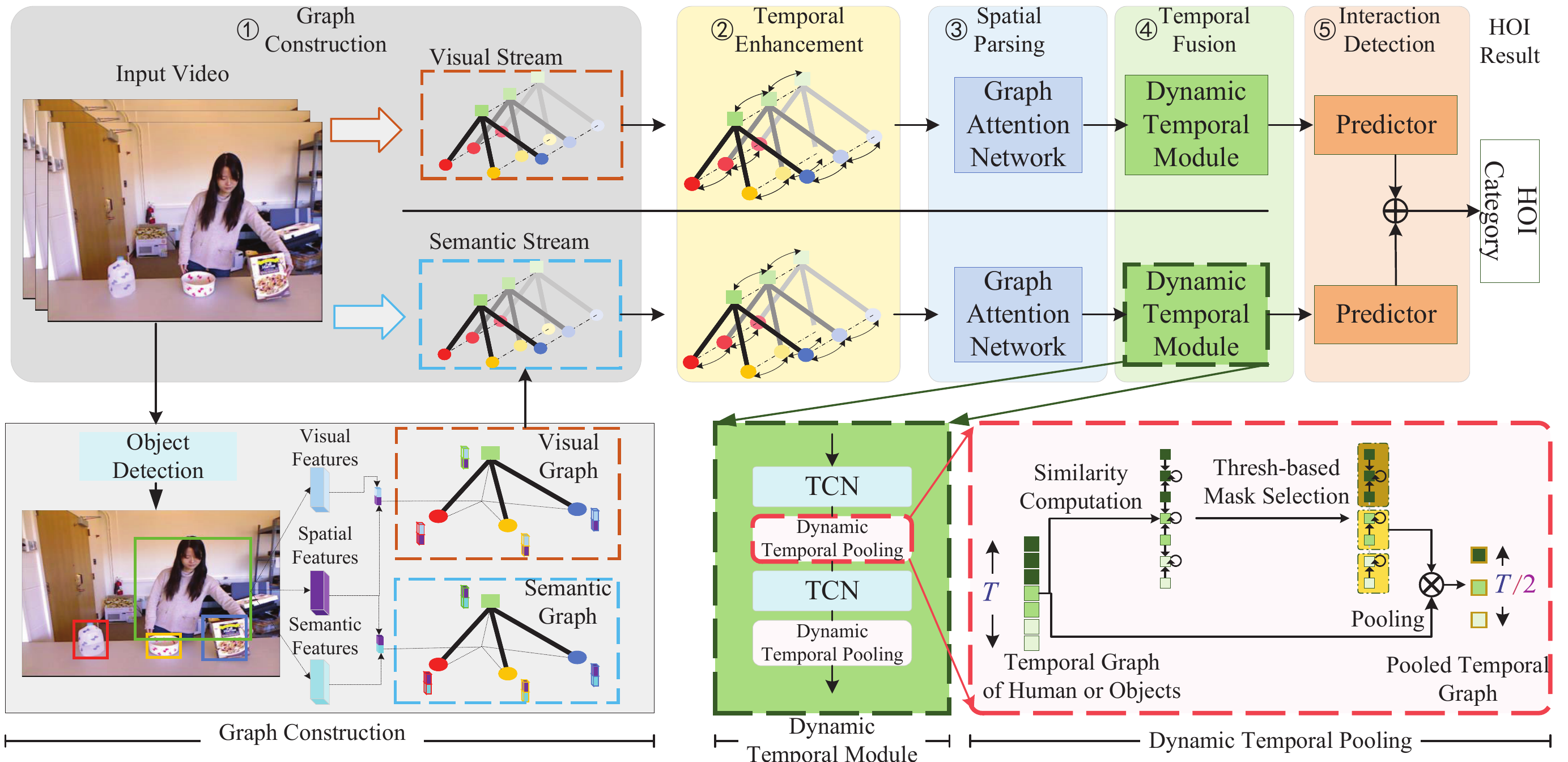}
	\caption{Framework of the proposed Spatial Parsing and Dynamic Temporal Pooling network. The Graph Attention Network parses the graph into sparse graph and merge the feature between nodes. The Dynamic Temporal Module learns keyframes from similarity and distinction between frames. }\label{TP}
\end{figure*}

In this section, we present the details of our Spatial Parsing and Dynamic Temporal Pooling network for HOI recognition in videos. 

\subsection{Preliminary}\label{graph}

Given an input video consisting of $ T $ frames such that the video includes $ N $ entities of either human or object class, we denote it as a graph $ G = \{V, E\}$, where the nodes $ V=\{v_1, v_2, \cdots, v_N\} $ correspond to $ N $ entities and $E$ denotes edges between nodes. 
We can define video features as $X=\{x^t_n\}$ and each individual is represented as a feature vector for $ n $-th entity at $ t $-th time step. 
Let $ A \in \mathbb{R}^{N \times N}$ be the simple adjacency matrix whose entry $ A_{ij} $ denotes whether node $ i $ and node $ j $ are connected. 


\subsection{The Proposed Learning Architecture}\label{arch}

The pipeline of our proposed method is shown in Fig. \ref{TP}, which can be divided into five stages, the first is the \textbf{Graph Construction}, the second is the \textbf{Temporal Enhancement}, the third is the \textbf{Spatial Parsing}, the fourth is the \textbf{Temporal Fusion}, and the last is the \textbf{Interaction Detection}. 
First of all, the initial is the \textbf{Graph Construction}, we build a three-dimensional video into a space-time graph by locating and extracting the multimodal features of each entity using pre-trained models. 
In the second step, before parsing the relationship between human-object or object-object in space, we added a step of \textbf{Temporal Enhancement}. This comes from our intuitive feeling that the relationship depends not only on the present moment but also on the past and the future. 
Combining information from the past and the future enables better exploration of spatial relationships in \textbf{Spatial Parsing}. 
Then after we get a more accurate spatial relationship, we do temporal parsing  in \textbf{Temporal Fusion} to focus on more meaningful frames. 
Finally, the category of HOI is predicted through \textbf{Interaction Detection}. The overall network structure is simple but effective and not simple stacking. 

\subsubsection{Graph Construction}
We use \textbf{Appearance Features}, \textbf{Spatial Features} and \textbf{Semantic Features} to construct initial graph for representing the whole video in this paper. By combining these three features, we get the node feature representation of the two graphs called \textit{visual graph} and \textit{semantic graph}. The adjacency matrix is initialized to be 1 for human-object edges and 0 for others. Moreover the edge $ A_{ij} $ between node $ i $ and node $ j $ represents the interaction between them, and our task is to learn it, the edge weight $ A_{i,j} $ between human and object is initialized to be 1 and 0 for others. During spatial graph parsing, the edge weight is dynamically adjusted.
\subsubsection{Temporal Enhancement}

Human and objects state transition indicate HOI actions. The node features extracted in the previous section are all based on the static images, and they reflect the nodes' states, e.g., attributes and relationships of humans and objects. However, the attribute and relationship transitions reflect the dynamic changes of objects over time and help detect the active objects in a crowded scene. Zhuo. et al. \cite{zhuo2019explainable} construct an explainable video action reasoning network using attributes and relations transitions. Compared with directly predicting HOI, this method has achieved better results, proving that the change of state overtime provides more effective clues for HOI recognition. However, it introduced attributes and relationship labels that were not in the original dataset, and the HOI categories obtained do not coincide with the original dataset. On the other hand, some attributes or relationships are challenging to define in language, which leads to inaccurate attributes and relationships that are explicitly represented. 

Unlike this method, an implicit method is adopted to extract the feature of state changes over time without introducing new annotations or definitions. Bi-directional Recurrent Neural Networks(BiRNN) is used to evolve the temporal state transition as illustrated in Fig.\ref{TP}. 
\subsubsection{Spatial Parsing}\label{ssubnet}

We denote the temporal enhanced graph in the video as $ G^t = (V^t, E^t)$. In $ G^t $, there are multiple objects, and not every object is related to the interaction represented in the current video: In the video of moving the cup, objects such as cups, microwave ovens, and bowls appear in the same scene, but only human interaction with the cup. Therefore, when identifying HOI categories, it is also essential to simultaneously identify the interacting human-object pairs. In this paper, we use the two-stream Graph Attention Network\cite{velivckovic2017graph} to parse the adjacency matrix and fuse the spatial features.

Graph Attention Network is a particular graph convolutional neural network based on attention\cite{vaswani2017attention}, and it does not need pre-defined edge weights between nodes but analyzes the relationship between nodes.
Formally, let $ \hat{x}^t_{v_i} $ and $ \hat{x}^t_{v_j} $ denote the temporal enhanced node feature os the source node and target node. First, the normalized relationship between the two nodes is mined using the attention.

\begin{equation}\label{transform}
a^t_{ij}=SoftMax(LeakyReLU(W^t_g[\hat{x}^t_{v_i}, \hat{x}^t_{v_j}]))
\end{equation}

Then the aggregation operation is performed as in graph convolution but by the attention scores.

\begin{equation}\label{gconv}
y^t_{v_i}=\sigma(\sum_{j \in \mathcal{N}}(a^t_{ij}W^t_h\hat{x}^t_{v_j}))
\end{equation}

We have two steam inputs, each of which embodies different modalities of this video. Using GAT, we can analyze the actual interactive objects in the video through the attention between nodes from two streams, thereby judging the interaction between the objects. 
Finally, the features extracted from the two-stream spatial networks are fed into the next temporal fusion networks.

\subsubsection{Temporal Fusion}\label{tsubnet}
Once the spatial features of each frame of the graph are extracted, we process the features of the graph in the temporal dimension to provide a feature panorama of the entire fragment.
As we know, in the frame sequence, not all the frames are of equal importance for the final classification, and the proposed model should pay more attention to the representative frames. Previous works always assigned different weights to different frames by attention mechanism. However, the weights of each frame are thoroughly learned by the network implicitly, and it cannot give a reasonable explanation why the model pays more attention to some dedicated frames. 

Our idea is to select a few frames from the frame sequence, and these frames must have sufficient discrimination and good quality. Therefore, we assign weights to frames based on their quality and similarity among frames, rather than letting the network learn by itself. This is the most significant difference between this model and other methods. As shown in the Temporal Fusion in Fig. \ref{TP}, this module contains a two-stream Dynamic Temporal Module(DTM), each DTM is alternately stacked by TCN and Dynamic Temporal Pooling(DTP). DTP is responsible for selecting the most appropriate frames within a window as keyframes. The details of DTP are described in Sec. \ref{dtp}.

\subsubsection{Interaction Detection}
Finally, for each node, the merged feature is fed into a readout function to output a label:	 

Given the node features from the parsed spatio-temporal graph as $ \hat{Y} = [\hat{y}_h, \hat{y}_{v_1}, \cdots, \hat{y}_{v_{N-1}}] $, the prediction of interaction categories can be written as:

\begin{equation}
H = R(\hat{y}_{h})
\end{equation}

where the readout function $ R(\cdot) $ computes output subactivity label $ A_h $ for human node feature $ \hat{y}_h $ and the readout functions $ R(\cdot) $ in this paper is a two fully connected layers MLP.

Nevertheless, for affordance prediction, we concatenate human node features along with object node features and feed them into affordance readout layer:

\begin{equation}
O_i = R(Concat[\hat{y}_{h}, \hat{y}_{v_i}])
\end{equation}

\textbf{Loss Function}
We use two standard cross-entropy loss $ \mathcal{L}_H $ and $ \mathcal{L}_O $ to constrain the two classifiers. The overall loss is a weighted sum of the two losses and can be written as:

\begin{equation}
	\mathcal{L} =\mathcal{L}_H +\lambda\mathcal{L}_O
\end{equation}


		

\subsection{Dynamic Temporal Pooling module (DTP)}\label{dtp}
This module aims to define a general, end-to-end strategy that allows one to merge temporal features hierarchically dynamically. Formally, the input video after spatial parsing $ Y \in \mathbb{R}^{C \times T}, Y=\{ y^1, y^2, \cdots\, y^T\} $ can be regarded as a sparse graph, and only the adjacent frames are connected. We seek to define a strategy to generate a new coarsened graph containing $ T^\prime < T $ nodes and produce more refined features. Existing methods usually use average pooling or max pooling along the time axis to merge the adjacent frames features. However, we seek to assign different importance to the adjacent frames. In other words, we need our model to learn a temporal fusion strategy that will generalize within segments with different numbers of frames, dynamically fuse redundant frames and select the most representative ones.

The critical point of our DTP is that it uses the similarity between adjacent frames and frames distinction as a mask for the frame selection. Such mechanisms can allow frames to spontaneously aggregate into multiple clusters, reducing node redundancy. Our pooling operation can be regarded as a local (in a small window), bottom-up node fusion method, and in the process of fusion, it is selective to merge instead of selecting all the frames.

\textbf{Similarity Computing.} 
We intend to focus more on frames with distinct features and reduce redundancy. Lower attention weights are assigned to those frames that share similar features, while higher attention weights are assigned to those that differ significantly. The attention weights of a frame depend on its relationship and differences with the other frames. Given a segment with $ \tau $ frame nodes $ Y^t = \{ y^{t - \tau}, \dots, y^t, \dots, y^{t + \tau}\} $, we use $ y^t $ as the center of pooling operations. The similarity score $ s^{ij} $ between node $ y^i$ and $y^j$, $i,j \in [t - \tau, t + \tau], i< j$ is calculated as follows:

\begin{equation}
s^{ij} = \sum_{k = i}^{j - 1} \theta(y^k)\phi(y^{k + 1})
\end{equation}

where $ \theta(y^k)=W_\theta y^k$ and $ \phi(y^{k + 1}) =W_\phi y^{k + 1}$ are two embeddings to map the feature to another space. By obtaining the similarity scores, we get the similarity between these three frames. The mask for node selection is calculated as follows:

\begin{equation}
    s^i = -\frac{1}{N}\sum_{j,i \neq j } s^{ij}
\end{equation}

where $ N $ is the number of frames in the window of $ y $. When a frame is more similar to other frames, its mask $ s^i $ is smaller, and the assigned weight is smaller.

\textbf{Distinction Computing}
In the entire video, the importance of the information contained in each frame is different. It is necessary to weaken the influence of the frame with low distinction and strengthen the influence of high-distinction frames. In time fusion, each frame should be assigned a critical weight. Our attention realization is calculated as follows.

\begin{equation}
 d^i = 	\psi(y^i)
\end{equation}

where $ \psi $ is an distinction score generator for assigning $ d^i $ for each frame. The more discriminative frame is assigned, the greater the weight.

The final $ Softmax(d^i + s^i) $ is the normalized mask. After getting the mask based on similarity and distinction, the pooling operation is processed as follows:

\begin{equation}
\hat{y} = \sum_{i}^{i \in [t - \tau, t + \tau]}Softmax(d^i + s^i) y^{i}
\end{equation}

As is shown in Dynamic Temporal Pooling in Fig. \ref{TP}, given an input video with 7 frames, we calculate the similarity score and do graph pooling with a stride of 2. Then we can get a video with 3 frame nodes, but the output is selective and weighted. From this figure, we can see that the assigned weights of these three frames are different.
\begin{table}
\centering
	\caption{Comparison of accuracies of spatial parsing with different settings. "TE" represents Temporal Enhancement operation.}\label{tab:ablation}
	\begin{tabular}{lccll}
		\hline
		Methos& TE & Full connected &Subactivity &Affordance\\
		\hline
		\hline
		GCN&\usym{2717}&\usym{2713}& 84.29&85.63 \\
		GCN&\usym{2717} &\usym{2717}&85.21&85.89 \\
		GAT&\usym{2717}&\usym{2717}& 85.98& 87.65\\
		
		\hline
		GCN&\usym{2713}&\usym{2713}& 87.99& 85.27\\
		GCN&\usym{2713} &\usym{2717}&88.75& 85.51\\
		
	    GAT&\usym{2713}&\usym{2717}& 91.83& 90.24\\
		\hline

		\hline
	\end{tabular}
\end{table}

\begin{table}
\centering
	\caption{Comparison of temporal fusion with different temporal module. "S" represents similarity metric and "D" represents distinction metric.}\label{tab:ablationoftemporal}
	\begin{tabular}{llcll}
	\hline
	Experiments&Models&Metrics&Subactivity &Affordance\\
		\hline
		\hline
		A&Baseline &-& 87.58& 87.56\\
		B&AvgP&-& 89.82& 89.26\\
		C&MaxP&-& 88.36& 89.81\\
		D&RNN&-&89.23& 88.60\\
		\hline
		E&DTP&S& 90.60& 89.27\\
		F&DTP&D& 88.87&89.45 \\
		G&DTP&S+D& 91.83& 90.24\\
		\hline
		\hline
	\end{tabular}
\end{table}

\section{Experiments}

\subsection{Datasets}

The CAD-120 dataset is a video HOI detection dataset containing 120 RGB-D videos of 4 subjects performing 10 indoor activities daily. Each activity is composed of a more detailed sub-action video clip. In each video clip, humans were annotated with activity labels from a set of 10 sub-activity categories (e.g., pour, move), and each object was annotated using a set of 12 affordance categories (e.g., pourable, movable) ) to annotate the affordance labels. The frame length of each clip ranges from 22 to 150 frames. 

The metrics used to evaluate our model on the HOI detection task on the CAD-120 dataset are i) the subactivity F1 score and ii) the object affordance F1 scores computed for the human subactivity classification and the object affordance classification. In this paper, we only additionally use bounding boxes of humans and objects and no other additional data.

The Something-Something V2 dataset is a medium-sized video action recognition dataset. The biggest difference between it and general datasets is that its content defines atomic actions, and this dataset pays particular attention to time-series relationships. The dataset contains 174 common human-object interactions. During production, the dataset producer chooses an action category (verb), performs it, and uploads a video with an arbitrary object (noun) accordingly. The dataset has a total of 12,554 different object descriptions.
Something-Else \cite{materzynska2020something} (built on \cite{goyal2017something} with the compositional setting forcing the combinations of action and objects cannot overlap between training and testing sets). Something-Else contains 174 categories of activities but only 112, 795 videos (54, 919 for training and 57, 876 for testing) with the compositional setting.

\subsection{Implementation Details}

We now discuss implementation details from both the model and training perspectives.

\textbf{Model}: We use uniform sampling to select a fixed number of frames, T, from each video clip (we use T=8 for our experiments on the CAD-120 dataset). Furthermore, we extract RoI crops from each frame and reshape them to a fixed size of 224×224×3. In our experiments, we use ResNet-50 as a feature extractor to generate features for each graph node. Meanwhile, we use a pre-trained ResNet to generate more generalized features. To integrate human and object localization information, we append the normalized bounding box coordinates of the human/object to their respective semantic node features and appearance node features.

\textbf{Training}: On the CAD-120 dataset, we implement our model using PyTorch. During training, we set $\lambda = 1 $ for the overall loss to balance the loss between activity classification and affordance classification. We use an SGD optimizer with an initial learning rate of $2e^{-5}$, a learning rate decay factor of 0.8, and a step decay of 20 epochs. We train our model on Nvidia RTX 2080Ti GPU for a total of 300 epochs.

Similar to the training process on the CAD-120 dataset, we train our models in 30 epochs and decay the learning rate by the factor of 10 for every 5 epochs on Something-Else dataset. All our models are fed with 16 frames and trained in a batch size of 72 on this dataset. Because of the lack of object type information, we mark all object types as "Object."

\subsection{Ablation Study}
In this section, we analyze and validate the role of each component of the proposed SPDTP network, on the CAD-120 dataset.

\subsubsection{Role of Temporal Enhancement}
In general, changes in attributes, in spatial positions, or in distance between instances provide cues for the parsing of spatial graph. To test this hypothesis, we designed an experiment that removed the temporal enhancement to verify its effectiveness. 

Experimental results are show in Table. \ref{tab:ablation}. There is a significant drop in the performance of the network, regardless of the spatial parsing variant. We argue that because of the lack of temporal clus, resulting in the spatial graph parsing difficulty increased, and ultimately affected the performance of the network.

\subsubsection{Role of Attention in spatial subnet}


In spatial subnet, we use GAT as our basic block, which utilizes an attention mechanism to learn the strength of the relations between nodes. Similarly, we also designed experiments to verify its effectiveness and use a basic graph convolution network(GCN) to replace the GAT in the spatial subnet. The difference between GCN and GAT is that the adjacent matrix of the former is fixed, while the latter can be dynamically generated by input data. Experiment results are shown in Table. \ref{tab:ablation}. We observe improvement while using GAT as a basic block instead of GCN. Not all objects in the scene are related to the current action. The fixed adjacency matrix causes the network to be affected by irrelevant features, thereby dropping the final performance of the network. Simultaneously, we can observe that the more predefined edges, the worse the performance of the network(the result of GCN is better than full connected GCN). Since the CAD-120 dataset mainly contains the interaction between humans and objects, there are almost no interactions between objects. 
\begin{figure}
    \centering
	\includegraphics[width=200px]{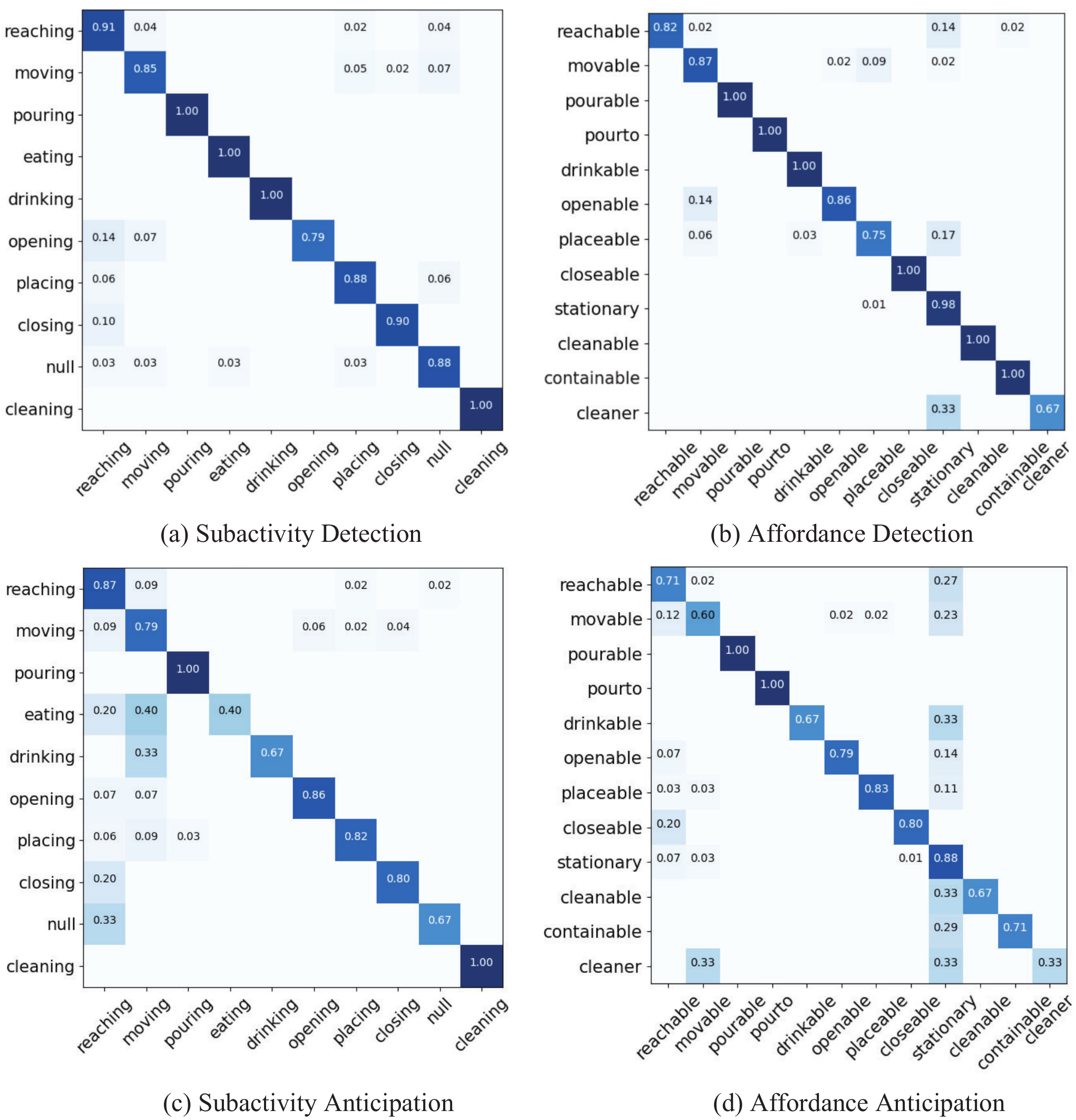}
	\caption{Confusion matrices for human-object interaction detection setting on CAD120 dataset(Subactivity in the left and affordance in the right). It can be seen that the two actions of opening and reaching are easily confused, and the video clips of the two actions are similar in appearance.}\label{fig:confusion}
\end{figure}

\subsubsection{Role of Dynamic Temporal Module}


Finally, we verify the role of the entire dynamic temporal pooling. We use temporal convolution without pooling as a baseline. Then we add average pooling or max pooling in the process of temporal convolutions, or we use RNN to replace temporal convolutions. Finally, we use the proposed DTP to replace the original pooling.

Experimental results are shown in Table \ref{tab:ablationoftemporal}. Compared with the baseline, the accuracy of the model with pooling has been improved to varying degrees, indicating that pooling has the ability to eliminate redundancy and extract salient features to a certain extent.
Compared with avg pooling and max pooling, our proposed DTP is improved by 2.01\% and 3.47\%, respectively.  This is because the proposed DTP can infer the importance of features from two aspects of \textbf{Similarity} and \textbf{Distinction}, which is more effective than avg pooling and max pooling. At the same time, it can be seen from the results that neither the lack of \textbf{Similarity} nor \textbf{Distinction} can achieve the best effect. 

\subsection{Quantitative Evaluation}

\textbf{Confusion Matrix:} The confusion matrices for detection and prediction tasks on CAD-120 are shown in Fig. \ref{fig:confusion}. The ordinate of the confusion matrix represents the ground-truth label, and the abscissa represents the predicted label. Each value in each row represents the probability that a category is predicted to be another category. From the confusion matrix of affordance detection, it can be seen that most of the wrong results are wrongly predicted to be static. The misrecognition is particularly pronounced in the 
category with the ground-truth label cleaner, as objects in this category tend to move relatively little during the sub-activity.

\begin{table}
\centering
	\caption{Comparison with the state-of-the-art on HOI detection task on CAD-120}\label{tab:sota}
	\begin{tabular}{lcc}
		\hline
		\multirow{3}{*}{Methods}&\multicolumn{2}{c}{F1 Scores}\\
		\cline{2-3}
		&Human &Object\\
		&Activity & Affordance\\
		\hline
		\hline
		ATCRF\cite{koppula2015anticipating} & 80.4& 81.5\\
		S-RNN\cite{jain2016structural}  & 83.2& 88.7\\
		S-RNN (multi-task)\cite{jain2016structural} & 82.4 & 91.1\\
		GPNN\cite{qi2018learning} & 88.9 & 88.8\\
		LIGHTEN w/o Seg-RNN \cite{sunkesula2020lighten}& 85.9 & 88.9 \\
		LIGHTEN (full model)\cite{sunkesula2020lighten} & 88.9 &\textbf{92.6}\\
		\hline
		SPDTP & \textbf{91.8}& 90.2\\
		\hline
		\hline
	\end{tabular}
\end{table}

\begin{table}
\centering
	\caption{Comparison with the state-of-the-art on anticipation task on CAD-120}\label{tab:anti}
	\begin{tabular}{lcc}
		\hline
		\multirow{3}{*}{Methods}&\multicolumn{2}{c}{F1 Scores}\\
		\cline{2-3}
		&Human &Object\\
		&Activity & Affordance\\
		\hline
		\hline
		ATCRF\cite{koppula2015anticipating} & 37.9& 36.7\\
		S-RNN\cite{jain2016structural}  & 62.3& 80.7\\
		S-RNN (multi-task)\cite{jain2016structural} & 65.6 & 80.9\\
		GPNN\cite{qi2018learning} & 75.6 & \textbf{81.9}\\
		LIGHTEN w/o Seg-RNN \cite{sunkesula2020lighten}& 73.2 & 77.6 \\
		LIGHTEN (full model)\cite{sunkesula2020lighten} & 76.4 &78.8\\
		\hline
		SPDTP & \textbf{81.9}& 78.8\\
		\hline
		\hline
	\end{tabular}
\end{table}

\subsection{Comparison with state-of-art}

Finally, we compare with previous results on CAD-120 in Table \ref{tab:sota} and \ref{tab:anti}. To our knowledge, all previous work on human-object interaction tasks in CAD-120 uses hand-crafted features provided by the CAD-120 dataset, except LIGHTEN\cite{sunkesula2020lighten}. We compare our method against the existing works on CAD-120: ATCRF\cite{koppula2015anticipating} , S-RNN\cite{jain2016structural} , and GPNN\cite{qi2018learning} and LIGHTEN\cite{sunkesula2020lighten}.  Our proposed network achieves state-of-the-art performance with subactivity detection F1 score of 91.75, but the affordance detection F1 score of 90.24 is lower than LIGHTEN(full model). This is because the Seg-RNN used in LIGHTEN(full model) gets three segments as input and leverages these inter-segment dependencies. It can significantly improve performance compared to predictions from frame-level temporal subnetworks. However, our network only inputs the current segment of video and gets better performance than LIGHTEN(w/\/\o Seg-RNN). Because the proposed DTP focuses  more on dynamic changes, we can achieve better results on Subactivity. However, affordance tends to be more static, resulting in inferior results from models such as S-RNN.  

Experimental results on Something-Else dataset are reported in Table \ref{tab:something-else}. "STRG" is short for Space-Time Region Graph. And "I3D +STIN +OIE + NL" means combining the appearance features from the I3D model and the features from the STIN+OIL+NL model by joint learning. "I3D, STIN+OIE+NL" means a simple ensemble model combining the separately trained I3D and the trained STIN+OIE+NL model. Different from other works in Table \ref{tab:something-else}, we use ResNet-50 as our appearance feature extractor. Experimental results show that our methods achieve state-of-the-art performance on both top-1 and top-5 accuracy. 

In particular, our "SPDTP" network gain 4.3\% and 5.8\% compared with "I3D + STIN + OIE + NL" on top-1 and top-5 accuracy, respectively. When in a ensemble combination way, out "I3D, SPDTP" model outperform the "I3D, STIN + OIE + NL" by 6.9\% and 6.0\%.

\subsection{Qualitative analysis on Spatial Parsed Graph}
\begin{figure}
	\includegraphics[width=250px]{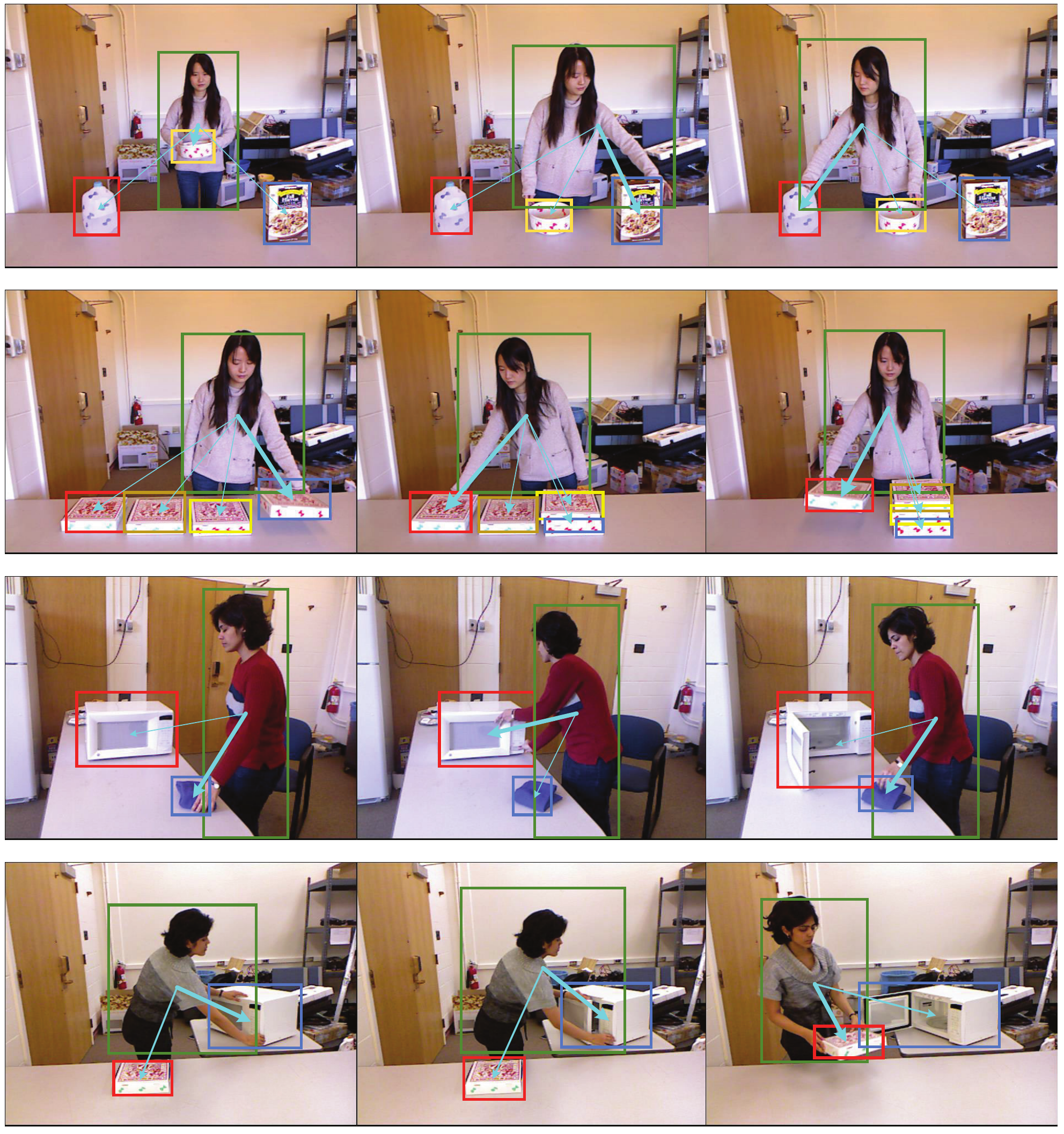}
	\caption{Visualization of some spatial parsed graphs. The thickness of the arrow represents the strength of the relationship.}\label{vis}
\end{figure}

Although we initialize the edges between humans and objects as 1, the SPDTP can automatically learn the relationship between humans and objects. Fig. \ref{vis} shows the spatial parsing results of some videos. The stronger the relationship, the thicker the arrow. As can be seen from the thickness of the arrows, the relationship between the proposed network parsed is correct. This proves the effectiveness of our network.

\subsection{Qualitative analysis on Temporal Parsed Graph}
\begin{figure*}
	\includegraphics[width=500px]{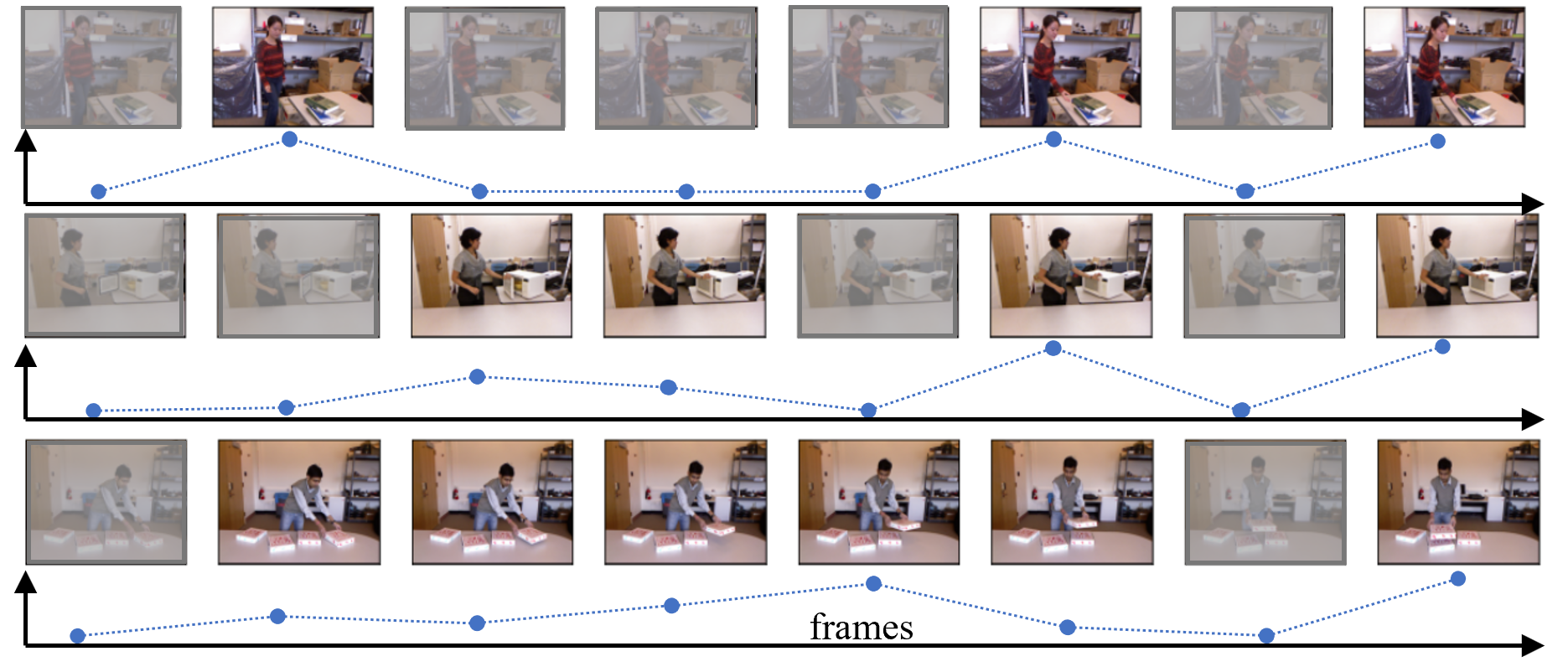}
	\caption{Visualization of temporal parsed graphs. The frames listed are those with a higher weight assigned.}\label{vist}
\end{figure*}

Fig. \ref{vist} shows the different weights assigned to different video frames by our proposed SPDTP in the process of detecting interactions. The height of the curve shows the size of the assigned weight. As can be seen from the figure, the network mainly focuses on three parts, one is the starting state of the action, the other is the rapid change in the middle, and the third is the final result state, which is consistent with our human understanding of interaction.

\begin{table}
\centering
	\caption{Comparison with the state-of-the-art on Something-Else.}\label{tab:something-else}
	\begin{tabular}{lcc}
		\hline
		\multirow{2}{*}{Method}&\multicolumn{2}{c}{Accuracy(\%)}\\
		\cline{2-3}
		&Top-1 &Top-5\\
		\hline
		\hline
		STIN\cite{materzynska2020something}& 47.1 &75.2\\
		STIN + OIE \cite{materzynska2020something}& 51.3& 79.3\\
		STIN + OIE + NL \cite{materzynska2020something}& 51.4& 79.3\\
		STRG \cite{wang2018videos}&52.3&78.3\\
		I3D\cite{shi2020skeleton} &46.8&72.2\\
		I3D + STIN + OIE + NL \cite{materzynska2020something}& 54.5& 79.4\\
		SPDTP & \textbf{58.8} & \textbf{84.2} \\
		\hline
		I3D, STIN + OIE + NL \cite{materzynska2020something}& 58.1 &83.2\\
		I3D, SPDTP & \textbf{65.0} & \textbf{89.2} \\
		\hline
		\hline
	\end{tabular}
\end{table}

\section{Conclusion}

In this paper, we proposed a hierarchical approach for identifying Human-Object Interaction. The spatial subnet parses the dense graph with a two-stream network and generates a sparse graph. And a temporal subnet with dynamic temporal pooling reduces the redundancy in time and extracts temporal features. 
Experimental results validate the effectiveness of our novel model components, and our proposed network achieves better performance of our proposed technique on two public video datasets.

\bibliographystyle{./IEEEtran}
\bibliography{./conference_101719}

\begin{thebibliography}{10}
\providecommand{\url}[1]{#1}
\csname url@samestyle\endcsname
\providecommand{\newblock}{\relax}
\providecommand{\bibinfo}[2]{#2}
\providecommand{\BIBentrySTDinterwordspacing}{\spaceskip=0pt\relax}
\providecommand{\BIBentryALTinterwordstretchfactor}{4}
\providecommand{\BIBentryALTinterwordspacing}{\spaceskip=\fontdimen2\font plus
\BIBentryALTinterwordstretchfactor\fontdimen3\font minus
  \fontdimen4\font\relax}
\providecommand{\BIBforeignlanguage}[2]{{%
\expandafter\ifx\csname l@#1\endcsname\relax
\typeout{** WARNING: IEEEtran.bst: No hyphenation pattern has been}%
\typeout{** loaded for the language `#1'. Using the pattern for}%
\typeout{** the default language instead.}%
\else
\language=\csname l@#1\endcsname
\fi
#2}}
\providecommand{\BIBdecl}{\relax}
\BIBdecl

\bibitem{ren2015faster}
S.~Ren, K.~He, R.~Girshick, and J.~Sun, ``Faster r-cnn: Towards real-time
  object detection with region proposal networks,'' \emph{arXiv preprint
  arXiv:1506.01497}, 2015.

\bibitem{wang2018repulsion}
X.~Wang, T.~Xiao, Y.~Jiang, S.~Shao, J.~Sun, and C.~Shen, ``Repulsion loss:
  Detecting pedestrians in a crowd,'' in \emph{Proceedings of the IEEE
  Conference on Computer Vision and Pattern Recognition}, 2018, pp. 7774--7783.

\bibitem{minderer2019unsupervised}
M.~Minderer, C.~Sun, R.~Villegas, F.~Cole, K.~Murphy, and H.~Lee,
  ``Unsupervised learning of object structure and dynamics from videos,''
  \emph{arXiv preprint arXiv:1906.07889}, 2019.

\bibitem{wang2019fast}
Q.~Wang, L.~Zhang, L.~Bertinetto, W.~Hu, and P.~H. Torr, ``Fast online object
  tracking and segmentation: A unifying approach,'' in \emph{Proceedings of the
  IEEE/CVF Conference on Computer Vision and Pattern Recognition}, 2019, pp.
  1328--1338.

\bibitem{bertinetto2016fully}
L.~Bertinetto, J.~Valmadre, J.~F. Henriques, A.~Vedaldi, and P.~H. Torr,
  ``Fully-convolutional siamese networks for object tracking,'' in
  \emph{European conference on computer vision}.\hskip 1em plus 0.5em minus
  0.4em\relax Springer, 2016, pp. 850--865.

\bibitem{dabral2018learning}
R.~Dabral, A.~Mundhada, U.~Kusupati, S.~Afaque, A.~Sharma, and A.~Jain,
  ``Learning 3d human pose from structure and motion,'' in \emph{Proceedings of
  the European Conference on Computer Vision (ECCV)}, 2018, pp. 668--683.

\bibitem{dabral2019multi}
R.~Dabral, N.~B. Gundavarapu, R.~Mitra, A.~Sharma, G.~Ramakrishnan, and
  A.~Jain, ``Multi-person 3d human pose estimation from monocular images,'' in
  \emph{2019 International Conference on 3D Vision (3DV)}.\hskip 1em plus 0.5em
  minus 0.4em\relax IEEE, 2019, pp. 405--414.

\bibitem{guler2018densepose}
R.~A. G{\"u}ler, N.~Neverova, and I.~Kokkinos, ``Densepose: Dense human pose
  estimation in the wild,'' in \emph{Proceedings of the IEEE conference on
  computer vision and pattern recognition}, 2018, pp. 7297--7306.

\bibitem{gupta2015visual}
S.~Gupta and J.~Malik, ``Visual semantic role labeling,'' \emph{arXiv preprint
  arXiv:1505.04474}, 2015.

\bibitem{zhuang2017care}
B.~Zhuang, Q.~Wu, C.~Shen, I.~Reid, and A.~v.~d. Hengel, ``Care about you:
  towards large-scale human-centric visual relationship detection,''
  \emph{arXiv preprint arXiv:1705.09892}, 2017.

\bibitem{gao2018ican}
C.~Gao, Y.~Zou, and J.-B. Huang, ``ican: Instance-centric attention network for
  human-object interaction detection,'' \emph{arXiv preprint arXiv:1808.10437},
  2018.

\bibitem{qi2018learning}
S.~Qi, W.~Wang, B.~Jia, J.~Shen, and S.-C. Zhu, ``Learning human-object
  interactions by graph parsing neural networks,'' in \emph{Proceedings of the
  European Conference on Computer Vision (ECCV)}, 2018, pp. 401--417.

\bibitem{chao2018learning}
Y.-W. Chao, Y.~Liu, X.~Liu, H.~Zeng, and J.~Deng, ``Learning to detect
  human-object interactions,'' in \emph{2018 ieee winter conference on
  applications of computer vision (wacv)}.\hskip 1em plus 0.5em minus
  0.4em\relax IEEE, 2018, pp. 381--389.

\bibitem{gibson2014ecological}
J.~J. Gibson, \emph{The ecological approach to visual perception: classic
  edition}.\hskip 1em plus 0.5em minus 0.4em\relax Psychology Press, 2014.

\bibitem{koppula2013learning}
H.~S. Koppula, R.~Gupta, and A.~Saxena, ``Learning human activities and object
  affordances from rgb-d videos,'' \emph{The International Journal of Robotics
  Research}, vol.~32, no.~8, pp. 951--970, 2013.

\bibitem{jain2016structural}
A.~Jain, A.~R. Zamir, S.~Savarese, and A.~Saxena, ``Structural-rnn: Deep
  learning on spatio-temporal graphs,'' in \emph{Proceedings of the ieee
  conference on computer vision and pattern recognition}, 2016, pp. 5308--5317.

\bibitem{chiou2021st}
M.-J. Chiou, C.-Y. Liao, L.-W. Wang, R.~Zimmermann, and J.~Feng, ``St-hoi: A
  spatial-temporal baseline for human-object interaction detection in videos,''
  in \emph{Proceedings of the 2021 Workshop on Intelligent Cross-Data Analysis
  and Retrieval}, 2021, pp. 9--17.

\bibitem{simonovsky2017dynamic}
M.~Simonovsky and N.~Komodakis, ``Dynamic edge-conditioned filters in
  convolutional neural networks on graphs,'' in \emph{Proceedings of the IEEE
  conference on computer vision and pattern recognition}, 2017, pp. 3693--3702.

\bibitem{yuan2017temporal}
Y.~Yuan, X.~Liang, X.~Wang, D.-Y. Yeung, and A.~Gupta, ``Temporal dynamic graph
  lstm for action-driven video object detection,'' in \emph{Proceedings of the
  IEEE international conference on computer vision}, 2017, pp. 1801--1810.

\bibitem{defferrard2016convolutional}
M.~Defferrard, X.~Bresson, and P.~Vandergheynst, ``Convolutional neural
  networks on graphs with fast localized spectral filtering,'' \emph{arXiv
  preprint arXiv:1606.09375}, 2016.

\bibitem{chen2017deeplab}
L.-C. Chen, G.~Papandreou, I.~Kokkinos, K.~Murphy, and A.~L. Yuille, ``Deeplab:
  Semantic image segmentation with deep convolutional nets, atrous convolution,
  and fully connected crfs,'' \emph{IEEE transactions on pattern analysis and
  machine intelligence}, vol.~40, no.~4, pp. 834--848, 2017.

\bibitem{chen2015learning}
L.-C. Chen, A.~Schwing, A.~Yuille, and R.~Urtasun, ``Learning deep structured
  models,'' in \emph{International Conference on Machine Learning}.\hskip 1em
  plus 0.5em minus 0.4em\relax PMLR, 2015, pp. 1785--1794.

\bibitem{liang2016semantic}
X.~Liang, X.~Shen, J.~Feng, L.~Lin, and S.~Yan, ``Semantic object parsing with
  graph lstm,'' in \emph{European Conference on Computer Vision}.\hskip 1em
  plus 0.5em minus 0.4em\relax Springer, 2016, pp. 125--143.

\bibitem{li2017situation}
R.~Li, M.~Tapaswi, R.~Liao, J.~Jia, R.~Urtasun, and S.~Fidler, ``Situation
  recognition with graph neural networks,'' in \emph{Proceedings of the IEEE
  International Conference on Computer Vision}, 2017, pp. 4173--4182.

\bibitem{marino2016more}
K.~Marino, R.~Salakhutdinov, and A.~Gupta, ``The more you know: Using knowledge
  graphs for image classification,'' \emph{arXiv preprint arXiv:1612.04844},
  2016.

\bibitem{xu2017scene}
D.~Xu, Y.~Zhu, C.~B. Choy, and L.~Fei-Fei, ``Scene graph generation by
  iterative message passing,'' in \emph{Proceedings of the IEEE conference on
  computer vision and pattern recognition}, 2017, pp. 5410--5419.

\bibitem{teney2017graph}
D.~Teney, L.~Liu, and A.~van Den~Hengel, ``Graph-structured representations for
  visual question answering,'' in \emph{Proceedings of the IEEE conference on
  computer vision and pattern recognition}, 2017, pp. 1--9.

\bibitem{liang2017interpretable}
X.~Liang, L.~Lin, X.~Shen, J.~Feng, S.~Yan, and E.~P. Xing, ``Interpretable
  structure-evolving lstm,'' in \emph{Proceedings of the IEEE conference on
  computer vision and pattern recognition}, 2017, pp. 1010--1019.

\bibitem{sunkesula2020lighten}
S.~P.~R. Sunkesula, R.~Dabral, and G.~Ramakrishnan, ``Lighten: Learning
  interactions with graph and hierarchical temporal networks for hoi in
  videos,'' in \emph{Proceedings of the 28th ACM International Conference on
  Multimedia}, 2020, pp. 691--699.

\bibitem{velivckovic2017graph}
P.~Veli{\v{c}}kovi{\'c}, G.~Cucurull, A.~Casanova, A.~Romero, P.~Lio, and
  Y.~Bengio, ``Graph attention networks,'' \emph{arXiv preprint
  arXiv:1710.10903}, 2017.

\bibitem{zhuo2019explainable}
T.~Zhuo, Z.~Cheng, P.~Zhang, Y.~Wong, and M.~Kankanhalli, ``Explainable video
  action reasoning via prior knowledge and state transitions,'' in
  \emph{Proceedings of the 27th ACM International Conference on Multimedia},
  2019, pp. 521--529.

\bibitem{vaswani2017attention}
A.~Vaswani, N.~Shazeer, N.~Parmar, J.~Uszkoreit, L.~Jones, A.~N. Gomez,
  L.~Kaiser, and I.~Polosukhin, ``Attention is all you need,'' \emph{arXiv
  preprint arXiv:1706.03762}, 2017.

\bibitem{materzynska2020something}
J.~Materzynska, T.~Xiao, R.~Herzig, H.~Xu, X.~Wang, and T.~Darrell,
  ``Something-else: Compositional action recognition with spatial-temporal
  interaction networks,'' in \emph{Proceedings of the IEEE/CVF Conference on
  Computer Vision and Pattern Recognition}, 2020, pp. 1049--1059.

\bibitem{goyal2017something}
R.~Goyal, S.~Ebrahimi~Kahou, V.~Michalski, J.~Materzynska, S.~Westphal, H.~Kim,
  V.~Haenel, I.~Fruend, P.~Yianilos, M.~Mueller-Freitag \emph{et~al.}, ``The"
  something something" video database for learning and evaluating visual common
  sense,'' in \emph{Proceedings of the IEEE International Conference on
  Computer Vision}, 2017, pp. 5842--5850.

\bibitem{koppula2015anticipating}
H.~S. Koppula and A.~Saxena, ``Anticipating human activities using object
  affordances for reactive robotic response,'' \emph{IEEE transactions on
  pattern analysis and machine intelligence}, vol.~38, no.~1, pp. 14--29, 2015.

\bibitem{wang2018videos}
X.~Wang and A.~Gupta, ``Videos as space-time region graphs,'' in
  \emph{Proceedings of the European conference on computer vision (ECCV)},
  2018, pp. 399--417.

\bibitem{shi2020skeleton}
L.~Shi, Y.~Zhang, J.~Cheng, and H.~Lu, ``Skeleton-based action recognition with
  multi-stream adaptive graph convolutional networks,'' \emph{IEEE Transactions
  on Image Processing}, vol.~29, pp. 9532--9545, 2020.

\end{thebibliography}
\vspace{12pt}
\color{red}

\end{document}